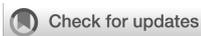

# Design and control of a low-cost non-backdrivable end-effector upper limb rehabilitation device

Fulan Li*†, Yunfei Guo†, Wenda Xu, Weide Zhang, Fangyun Zhao, Baiyu Wang, Huaguang Du and Chengkun Zhang

Futronics (NA) Corporation, Pasadena, CA, United States

This paper presents GARD, an upper limb end-effector rehabilitation device developed for stroke patients. GARD offers assistance force along or towards a 2D trajectory during physical therapy sessions. GARD employs a non-backdrivable mechanism with novel motor velocity-control-based algorithms, which offers superior control precision and stability. To our knowledge, this innovative technical route has not been previously explored in rehabilitation robotics. In alignment with the new design, GARD features two novel control algorithms: Implicit Euler Velocity Control (IEVC) algorithm and a generalized impedance control algorithm. These algorithms achieve $O(n)$ runtime complexity for any arbitrary trajectory. The system has demonstrated a mean absolute error of 0.023 mm in trajectory-following tasks and 0.14 mm in trajectory-restricted free moving tasks. The proposed upper limb rehabilitation device offers all the functionalities of existing commercial devices with superior performance. Additionally, GARD provides unique functionalities such as area-restricted free moving and dynamic Motion Restriction Map interaction. This device holds strong potential for widespread clinical use, potentially improving rehabilitation outcomes for stroke patients.

## 1 Introduction

Every forty seconds, an individual in the United States experiences a stroke, and every four minutes, a stroke leads to death. Approximately 7.6 million Americans aged 20 years and older have experienced stroke. As age progresses, the prevalence of stroke increases in both males and females. By 2030, the prevalence rate is projected to increase to 3.9% (1), making stroke a major health concern in the US. Research shows that between 30% to 66% of hemiplegic stroke patients experience limited arm motor function six months after a stroke, with only 5% to 20% demonstrating complete functional recovery (2). These low patient recovery rates are linked to diminished quality of life and increased risks to overall well-being.

Robotic-assisted therapy (RAT) has been found to be a valuable adjunct to conventional physical therapy for post-stroke upper extremity rehabilitation, particularly for subacute stroke patients (3). Compared to conventional physical therapy, RAT allows patients to undergo consistent and repetitive rehabilitation exercises, showing comparable and better outcomes (4). Among those frequently researched RAT devices (e.g., rehabilitation exoskeletons and end-effector devices), the end-effector rehabilitation devices are particularly popular due to their portability and affordability.

Meanwhile, the end-effector devices can still deliver patient outcomes that are comparable to more complex systems such as exoskeleton robots (4). The present article introduces the design and control of a low-cost end-effector rehabilitation device. The proposed device and control method shows the potential to offer tailored and long-term adaptive training for stroke patients and extend the device's product life span.

This paper introduces the Gantry Arm Rehabilitation Device (GARD), an accurate, low-cost, powerful end-effector-based upper limb rehabilitation robot, as shown in Figure 1E. GARD's design features a non-backdrivable mechanism, which enhances control accuracy and cost-effectiveness—an approach rarely explored in previous research. The device uses two ball-screw-driven linear actuators for movement along the $x$ and $y$ axes, providing superior stability along the $z$-axis compared to multi-link robots. Additionally, the direct-drive ball-screw mechanism offers better wear resistance and higher accuracy than belt-driven systems. Our device matches the functionalities of existing commercial robots, offering superior accuracy. Additionally, GARD provides unique features, such as area-restricted free moving, which enhances the flexibility of training protocols. Three main challenges are addressed in this paper:

First, a non-backdrivable mechanism presents challenges in constructing an accurate dynamics model. For instance, friction within the transmission can be significantly influenced by various factors such as load, temperature, configuration, lubrication levels, and velocity. To solve this, we develop our control methods based on the motor velocity control and a virtual dynamics model. This design dynamically compensates for elusive forces using the PID velocity control and calculates the desired velocity using a virtual dynamics model.

Second, the velocity-control-based trajectory-following method can lead to accumulating positional errors, causing the end-effector to deviate from the desired path. This error arises from the control discretization and physical limitations on the motor acceleration. Increasing the control frequency or using more powerful motors will reduce the error but cannot eliminate it. To address this, we develop the Implicit Euler Velocity Control (IEVC) algorithm, discussed in Section 5.1.2.

Third, trajectory tracking requires customization for individual users. Previous research and commercial products often focus on predefined trajectories with symbolic solutions (7, 9, 10). While assistive force calculations for symbolic trajectories are straightforward, generalizing on complex, customized trajectories requires additional fitting and tuning. We propose a novel control method that efficiently calculates assistive forces on a 2D discretized numerical trajectory. Unlike symbolic methods, this method handles any trajectory in a 2D workspace without fitting and tuning. With computational time unaffected by trajectory complexity, our method uniquely supports real-time customization.

## 2 Related works

End-effector upper limb rehabilitation devices refer to planar robots enabling movement along a 2D plane. Numerous studies have investigated rehabilitation techniques using these 2D planar robots through rigorous experimental methodologies (6–8, 11–14). There are three primary types of end-effector rehabilitation robots: multi-link robots (6, 11–13), pulley-driven robots (7, 14), and linear actuated robots (8) (Figure 1). The MIT MANUS is a well-known example of a multi-link rotary robot with a two-decade history (6). Stability of the vertical force, resulting from the user placing their arm on the device, has been a notable concern for multi-link rotary robots. Despite their portable and backdrivable direct drive mechanisms providing high accuracy, maintaining stability of the vertical force remains challenging (7). Researchers proposed pulley-driven robots and linear actuated robots to improve stability (15, 16). Pulley-drive robots attempted to employ a differential drive with pulleys and belts, to address the stability issues and improve dynamics calculation (7, 10, 14, 17). However, the pulley-drive robots require high maintenance due to the frequent wear and tear on the belt drives. To our knowledge, only one commercialized device claimed to solve these reported issues using a backdrivable linear actuated mechanism, but with minimum information revealed (8, 9). The current paper explores an alternative system architecture with a non-backdrivable mechanism to provide less maintenance than pulley-driven robots.

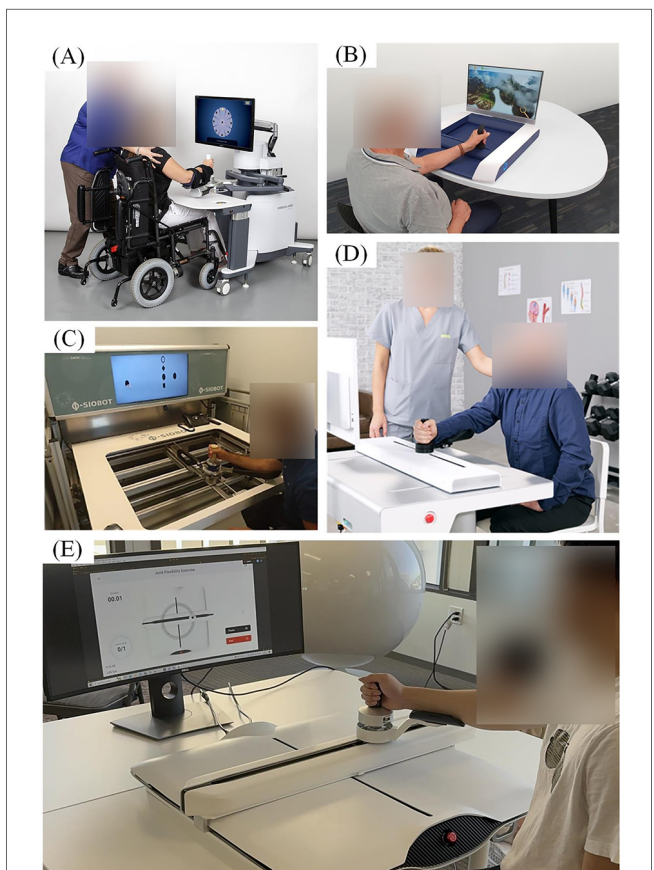

FIGURE 1
End-effector rehabilitation devices. (A) InMotion (5) arm rehabilitation device, commercialized based on the multi-link MIT MANUS (6) design. (B) H-MAN (7) arm rehabilitation device, utilizing an H-shaped differential pulley and belt mechanism. (C) Physiobot arm rehabilitation device using a backdrivable linear actuated mechanism (8). (D) ArmMotus M2 Gen (9) rehabilitation device utilizing a backdrivable linear actuated mechanism. (E) Our Gantry Arm Rehabilitation Device (GARD) operates in Assist-As-Needed Mode along a circular trajectory. The GARD can provide either assistance or resistance to support physical therapy training.

End-effector upper limb rehabilitation devices typically offer three training modes tailored to specific rehabilitation interventions: Powered Mode, Assist-As-Needed Mode, and Transparent Mode. These training modes aim to map the sequential stages of conventional physical therapy progression (14, 18, 19). The Powered Mode is primarily utilized in the early stages of rehabilitation for stretching and passive range of motion (ROM) exercises (20). Patients follow a predetermined 2D trajectory operated by the device, without using strength or exerting forces. The Assist-As-Needed Mode attempts to facilitate active range of motion (ROM) exercises and resistance training (20, 21). In this mode, the device provides assistance or resistance as needed. The Transparent Mode allows users unrestricted movement within the 2D workspace, facilitating engagement with rehabilitation games, promoting sensorimotor training, and enhancing tactile feedback, motor planning, and sensory and proprioceptive awareness (22).

# 3 Gantry arm rehabilitation device

## 3.1 Mechanical design

The Gantry Arm Rehabilitation Device (GARD) end-effector offers a flexible range of motion within a $65 \times 55$ cm 2D workspace. This device utilizes two rotary actuators coupled with ball screws, enabling linear movement along the $x$ and $y$ axis, as depicted in Figure 2. The ball screws operate at a ratio of 1 turn to 2 mm. The actuators can achieve a maximum speed of 80 revolutions per second, with acceleration limited to 800 revolutions per second squared. Additionally, each actuator provides a maximum torque output of 0.22 Nm.

## 3.2 Electrical design

The device uses a medical-grade power supply with a rated output of 24 V and 6.5 A, distributed via a power board to components such as the motor controller, force sensor, and controller board. Two motor controllers are connected to Brushless DC motors with 1000CPR optical encoders, enabling precise control of the $x$ and $y$-axis ball screws. Additionally, the system employs two limit switches for motor-homing purposes. For force sensing at the end-effector, a low-cost Galoce GPB160 3-axis load cell is employed, utilizing an ADS1232 SPI ADC converter. To manage low-level control and communication with the high-level controller, the limit switches, ADC converters, and motor drivers are all interfaced with an STM32 microcontroller unit (MCU).

## 3.3 Control hierarchy

GARD uses a two-level control hierarchy consisting of a High-level Virtual Dynamic Model and a Low-level Firmware Controller, as depicted in Figure 3A. The High-level Virtual Dynamic Model calculates the desired motor velocity from sensor data and user settings. The Low-level Firmware Controller handles data collection, timing, and motor control.

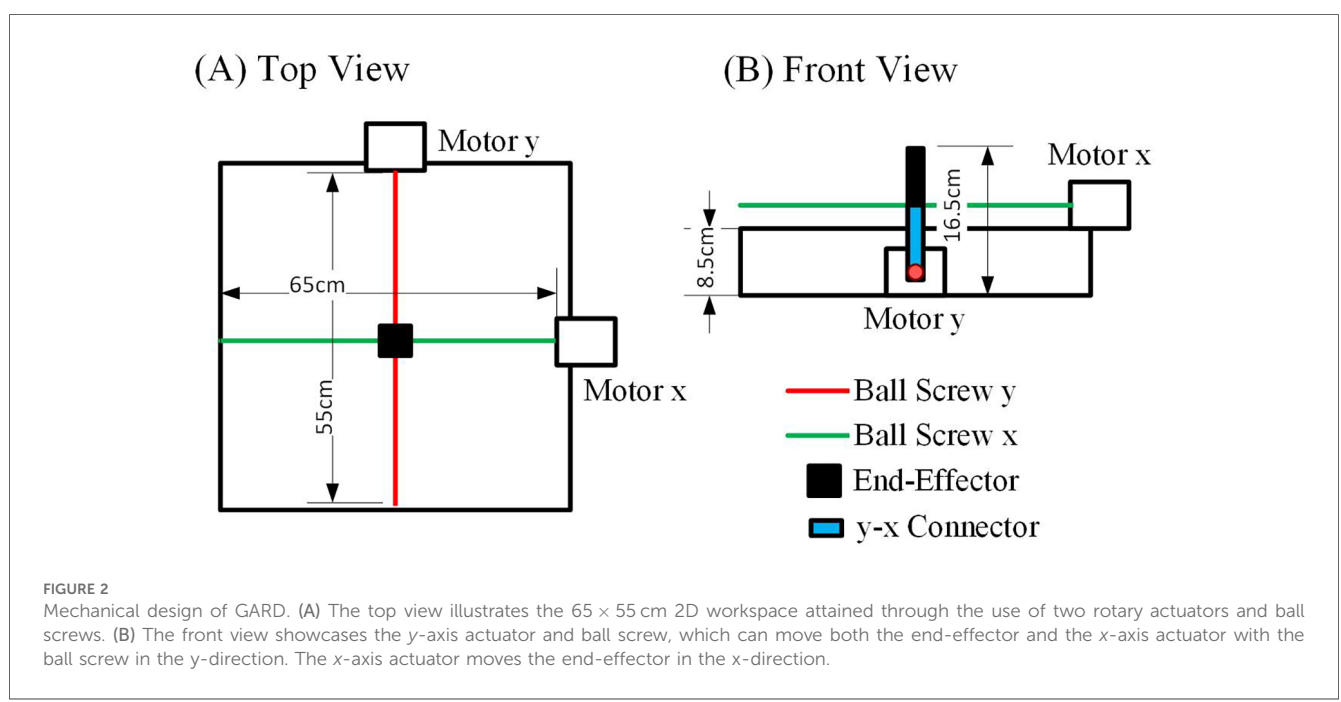

FIGURE 2
Mechanical design of GARD. (A) The top view illustrates the $65 \times 55$ cm 2D workspace attained through the use of two rotary actuators and ball screws. (B) The front view showcases the $y$-axis actuator and ball screw, which can move both the end-effector and the $x$-axis actuator with the ball screw in the $y$-direction. The $x$-axis actuator moves the end-effector in the $x$-direction.

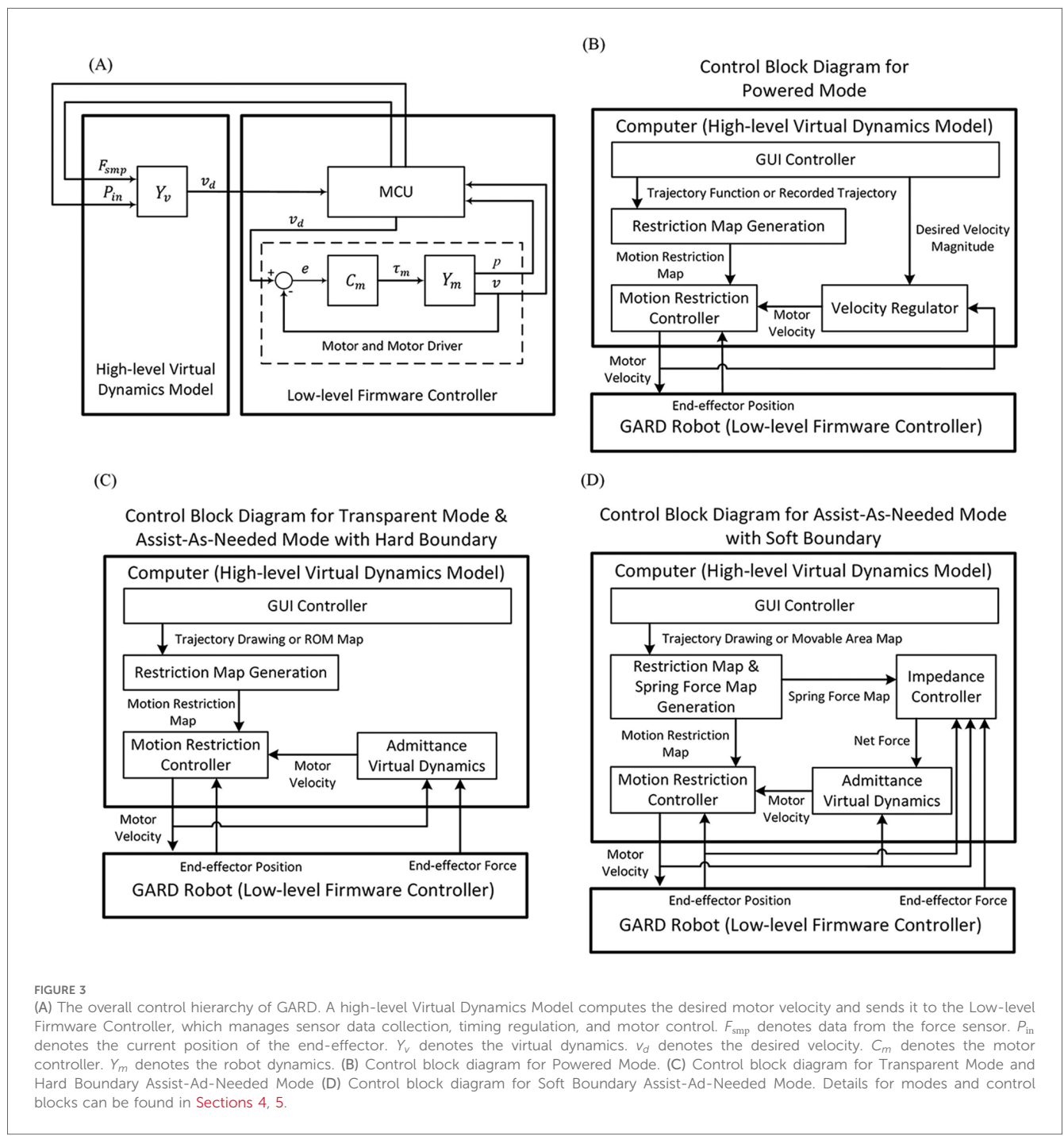

FIGURE 3
(A) The overall control hierarchy of GARD. A high-level Virtual Dynamics Model computes the desired motor velocity and sends it to the Low-level Firmware Controller, which manages sensor data collection, timing regulation, and motor control. $F_{smp}$ denotes data from the force sensor. $P_{in}$ denotes the current position of the end-effector. $Y_v$ denotes the virtual dynamics. $v_d$ denotes the desired velocity. $C_m$ denotes the motor controller. $Y_m$ denotes the robot dynamics. (B) Control block diagram for Powered Mode. (C) Control block diagram for Transparent Mode and Hard Boundary Assist-Ad-Needed Mode (D) Control block diagram for Soft Boundary Assist-Ad-Needed Mode. Details for modes and control blocks can be found in Sections 4, 5.

The current structural setup diverges from conventional admittance control systems. In conventional systems, the virtual dynamics model and motor control cannot be separated due to the influence of applied forces on robot dynamics. However, the GARD features a non-backdrivable end-effector, where user force does not influence end-effector motion, allowing the separation of the virtual dynamics model and the motor control. The separated structure allows us to establish a clear control hierarchy and maximize the integration advantages of commercial motor drivers. The integrated commercial motor driver firmware is more stable and efficient than the user's implementation on a multitask MCU because they are dedicated to motor velocity control.

## 4 GARD operation modes

Three distinct operation modes have been implemented to support patients at different stages of recovery: Powered Mode, Transparent Mode, and Assist-As-Needed Mode. A Graphical User Interface (GUI) is provided for all operational modes, offering GUI functionality and rehabilitation gaming features.

This section introduces the operation modes and their corresponding control algorithms.

## 4.1 Powered mode

Powered Mode is designed for passive rehabilitation, where the robot follows a pre-set trajectory at a user-defined speed. In this mode, exerted force on the end-effector does not influence the robot's motion. It is typically used during the early stages of recovery when external guidance is needed to complete arm movements.

The Powered Mode control hierarchy is shown in Figure 3B. The GUI controller collects user settings and inputs, sending them to the Motion Restriction Map Generator and the Velocity Regulator. The Velocity Regulator outputs a velocity vector based on the desired speed and current velocity direction. The Motion Restriction Controller adjusts the velocity vector to keep the end-effector within the desired moving area. We propose the IEVC algorithm that implements the Motion Restriction Controller function. The Motion Restriction Controller and IEVC are reused in all modes. More details are provided in Section 5.1.

## 4.2 Transparent mode

Transparent Mode is an interactive mode used in the later stages of recovery, supporting rehabilitation games and sensorimotor training. The Transparent Mode allows the user to move the end-effector within the user's Range of Motion (ROM). ROM is the area where users can safely move their arms without injuries. The interaction with the ROM boundary resembles encountering a hard-smooth-wall-like boundary.

Figure 3C illustrates the control hierarchy for Transparent Mode. The Admittance Virtual Dynamics block simulates the behavior of a mass on a rough surface, with friction and damping effects. Further details about this system can be found in Section 5.2.

## 4.3 Assist-As-Needed modes

Assist-As-Needed Modes are used for both active Range of Motion (ROM) and resistance training. Similar to the Transparent Mode, this mode allows free movement of the end-effector within the training area. However, the Assist-As-Needed Mode focuses on trajectory-based training, whereas Transparent Mode focuses on area-based training. There are two versions of this mode: hard boundary and soft boundary.

In the Assist-As-Needed Mode with hard boundary, the user can move the end-effector freely along a fixed trajectory. The user is restricted from deviating off the path similar to following a rail. The control structure of this mode is identical to the Transparent Mode. This is because a trajectory restriction and an area restriction are identical under our proposed algorithm.

The soft boundary version of Assist-As-Needed Mode is similar to the hard boundary version but allows the user to deviate from the path. The device applies automatic assistance force to guide the user back to the trajectory. The control hierarchy of the Assist-As-Needed Mode with soft boundary is outlined in Figure 3D. The Impedance Controller block calculates the assistance force for trajectory realignment. Details of the Impedance Controller can be found in Section 5.3.

## 5 Control algorithms

This section covers the control algorithms that enable GARD's operation modes. Section 5.1 explains the motion restriction and trajectory-following algorithms, Section 5.2 covers the simulated virtual dynamics, and Section 5.3 introduces the automatic assistance force calculation algorithms for soft boundary Assist-As-Needed Mode.

## 5.1 Motion restriction algorithms

**Truncation error accumulation:** A common practice in implementing the trajectory-following algorithm is to decompose the current velocity into tangential and radial velocity and adjust them accordingly. However, these types of algorithms share a common issue: the position error can accumulate over time, resulting in increased deviation and undefined positions. Take circular trajectory as an example: a trivial way to follow a circular trace is to dynamically calculate the tangential direction at the current position and march only in that direction. However, with this method in practice, the error caused by the control discretization and the motor acceleration limitations will accumulate, and will eventually result in an outward spiral trace instead of a circular trace. This example of truncation error accumulation is the same as the local truncation error accumulation in solving the initial value problems using the Explicit Euler Integral (23).

We propose the Implicit Euler velocity control (IEVC) algorithm to address the error accumulation issue during free moving along a trajectory restriction. This algorithm predicts the end-effector's next position and adjusts the current velocity, inspired by the Implicit Euler Method (23). However, we discover that this method can solve more than trajectory restriction: it can be naturally extended to 2D area restriction or even 3D spatial restriction. Hence, we define the **Motion Restriction Control** as the end-effector moving control under a virtual positional restriction. Areas that the end-effector is allowed to move are the **permitted area**, otherwise are the **prohibited area**. A **Motion Restriction Map** is a 2D matrix that can be viewed as a map marking the permitted and prohibited areas.

The proposed IEVC algorithm can effectively restrict the end-effector inside any permitted area, used in Figure 3 as the Motion Restriction Controller block. The IEVC allows the end-effector to move freely inside the permitted area and interact with a hard-smooth-wall-like area boundary. The following two subsections discuss the details of the Motion Restriction Map and IEVC algorithm.

### 5.1.1 Motion restriction map generation

The task space of GARD is discretized into $W_g \times H_g$ positions, forming a 2D matrix of equal size referred to as the Motion Restriction Map. The $\mathbf{M}_{rs}$ can be visualized as a grayscale image. Each pixel entry on the Motion Restriction Map corresponds to a physical GARD end-effector location. A pixel storing a value of 0 indicates that the corresponding location is the **prohibited area**; Any non-zero value indicates the corresponding location is the **permitted area**. The Motion Restriction Map will be initialized with all 0, meaning all prohibited areas.

For a trajectory restriction defined using implicit functions $f(x, y) = 0$ or more generalized area restriction defined using $\mathbf{F}(x, y) < \mathbf{0}$, the Motion Restriction Map can be generated by iterating through all entries $[i, j]$ and evaluating $abs(f(j, i)) < E_s$ or $\mathbf{F}(j, i) < \mathbf{0}$. $E_s$ is the trajectory width for trajectory restriction. If position $[i, j]$ meets the condition, the matrix entry will be assigned 1, marking it as a permitted area.

For other general user-defined motion restrictions such as a hand-drawn trajectory restriction, $\mathbf{M}_{rs}$ can be generated or edited using image editing tools.

### 5.1.2 Motion restriction controller (IEVC algorithm)

In simple terms, the IEVC algorithm restricts motion within the permitted area by predicting all possible positions for the next time step and selecting the optimal one. The optimal position will be set as the chasing target. There are three requirements for the chasing target:

1. The chasing target needs to be at a suitable distance according to the current speed.
2. The chasing target needs to be inside the permitted area.
3. The chasing target needs to be as aligned with the current velocity direction as possible.

The first requirement accounts for the fact that a higher end-effector speed means it will be farther in the next time step. The second requirement ensures the end-effector is always directed toward the permitted area. The third requirement filters the best candidate and allows the algorithm to be generalized into the area-based motion restriction.

Figure 4 illustrates an example of IEVC output calculation. The algorithm will first draw a circle centered at the current position $\mathbf{P}_{in}$ with a radius proportional to the current speed magnitude ($r = G_v * \|\mathbf{V}_{in}\|$). All positions on this circle satisfy requirement 1. The proportional gain $G_v$ is calculated by the transmission ratio between the motor and the end-effector. The algorithm then identifies intersection points between the circle and the permitted area, marking them as candidates. In the figure, 2 candidate points are marked with blue dots, both of them satisfy requirement 1 and 2. Then, the algorithm examines all candidates and chooses the one with the most aligned velocity as the chasing target. This target will satisfy all three requirements. Finally, the algorithm generates the final output velocity $\mathbf{V}_r$ using the chasing target direction and the projection of current velocity as the magnitude. The projection process simulates the momentum loss on collision with the wall, creating a hard-smooth-wall-like interaction with the area boundary. The example in Figure 4 also demonstrates that even when $\mathbf{P}_{in}$ is outside the permitted area (trajectory), the error will not accumulate under IEVC. The divergence distance will decay exponentially over time steps.

Algorithm 1 present the sudo code for IEVC. $\mathbf{P}_{in}$ denotes current position, $\mathbf{V}_{in}$ denotes current velocity, $\mathbf{M}_{rs}$ denotes the motion restriction map, $G_v$ denotes the transmission gain. The helper function sudo codes can be found in Algorithm 4.

## 5.2 Admittance virtual dynamics

The Admittance Virtual Dynamics block is used in both Transparent Mode and Assist-As-Needed Mode. This control block simulates adjustable mass-friction-damper dynamics, allowing the natural motion of the end-effector.

Given a simulated mass, we have the following equation of motion in the Laplace domain:

$$\begin{aligned} f &= s m_v v k_r \\ \frac{v}{f} &= \frac{1}{m_v k_r} \cdot \frac{1}{s} \end{aligned} \quad (1)$$

where, $m_v$ denotes the virtual mass. $k_r$ denotes the transmission ratio between the motor and the end-effector velocity. $T_s$ denotes the time step size.

The Laplacian domain model cannot be directly used for discrete control. There are two steps to convert the Laplacian domain model into discrete control. First, take the Z-transform with the Tustin estimator on Equation 1 gives:

$$\begin{aligned} \frac{v}{f} &= \frac{1}{m_v k_r} \cdot \frac{T_s}{2} \frac{z+1}{z-1} \\ \Rightarrow v &= \frac{T_s}{m_v k_r} \cdot \frac{f + z^{-1} f}{2} + z^{-1} v \end{aligned} \quad (2)$$

Second, take the inverse Z-transform on Equation 2 gives the time domain equation which can be used for discrete control:

$$v_k = \frac{T_s}{m_v k_r} \cdot \frac{f + f_{k-1}}{2} + v_{k-1} \quad (3)$$

Algorithm 2 shows the implementation of Equation 3. The algorithm takes current input force $\mathbf{F}_{in}$, input force on last time step $\mathbf{F}_{last}$, velocity on last time step $\mathbf{V}_{last}$, virtual mass $m_v$, damping ratio $\zeta_a$, and friction coefficient $\mu$ as input, and will output the desired velocity $\mathbf{V}_d$. The $\mathbf{V}_d$ will be sent to the Motion Restriction Controller. The motor velocity from the Motion Restriction Controller will be sent to low-level GARD firmware (e.g., MCU, motor controller) for close-loop velocity control.

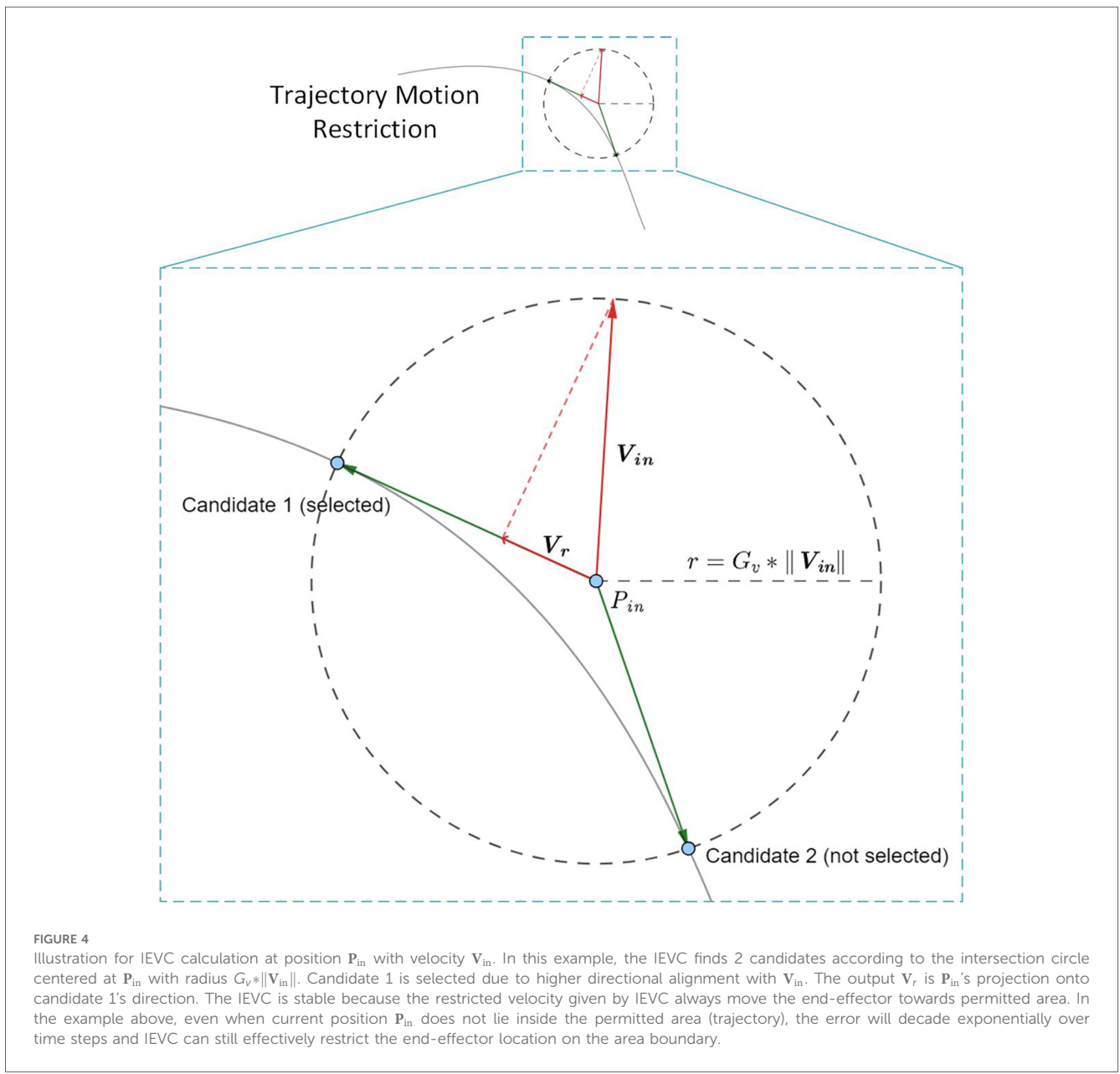

**FIGURE 4**
Illustration for IEVC calculation at position $\mathbf{P}_{in}$ with velocity $\mathbf{V}_{in}$. In this example, the IEVC finds 2 candidates according to the intersection circle centered at $\mathbf{P}_{in}$ with radius $G_v*\|\mathbf{V}_{in}\|$. Candidate 1 is selected due to higher directional alignment with $\mathbf{V}_{in}$. The output $\mathbf{V}_r$ is $\mathbf{P}_{in}$'s projection onto candidate 1's direction. The IEVC is stable because the restricted velocity given by IEVC always move the end-effector towards permitted area. In the example above, even when current position $\mathbf{P}_{in}$ does not lie inside the permitted area (trajectory), the error will decade exponentially over time steps and IEVC can still effectively restrict the end-effector location on the area boundary.

## 5.3 Impedance controller

The Impedance controller calculates assistive forces based on the current position and the motion restriction map. It operates in conjunction with the Admittance Virtual Dynamics in the soft boundary Assist-As-Needed Mode, simulating a spring-damper (impedance) boundary for the permitted area.

The output from the Impedance Controller is a composite force vector that combines: the end-effector sensor force, the simulated impedance spring force (referred to as spring force hereafter), and the simulated impedance damper force (referred to as damper force hereafter).

We propose a novel algorithm that efficiently calculates the impedance force on a Motion Restriction Map in real-time. This method provides a general solution for the discretized workspace, meaning that the complexity of the Motion Restriction Map has no impact on the computational cost. The algorithm works by precalculating a special map called the Spring Force Map using 2D convolution. The Spring Force Map allows the efficient calculation of impedance force at any given position in $O(1)$ time.

### 5.3.1 A simplified 1D impedance force calculation example

To explain the need for convolution, consider a simplified impedance force calculation problem in 1D. As illustrated in Figure 5A and Figure 5B, $M_{rs}(x)$ is our example 1D Motion Restriction Map. Each $x$ value corresponds to an end-effector location on the $x$ axis, meaning all positions between $2 \leq x \leq 6$ are permitted area. $Kernal_{sw}(x)$ is a unit impulse function and is used as the convolution kernel. The convolution result, seen in

**Algorithm 1** Implicit Euler velocity control (IEVC).

```
Input: P_in, V_in, M_rs, G_v
Output: V_r
 1:  R_circle ⇐ round(G_v * norm(V_in))
 2:  candidatesList ⇐ findIntersec(M_rs, R_circle, P_in)
 3:  if length(candidatesList) > 0 then
 4:    D_next ⇐ None
 5:    maxProj ⇐ −∞
 6:    for all P_c in candidatesList do
 7:      D_c ⇐ normalize(P_c − P_in)
 8:      curProj ⇐ (V_in · D_c)
 9:      if curProj > maxProj then
10:        maxProj ⇐ curProj
11:        D_next ⇐ D_c
12:      end if
13:    end for
14:    V_r ⇐ max(maxProj, 0) * D_next
15:    return V_r
16:  end if
```

**Algorithm 2** Algorithm 2 Admittance virtual dynamics.

```
Input: F_in, F_last, V_last, ζ_a, m_v, μ
Output: V_d
 1:  Y_m ⇐ T_s/m_v/k_r
 2:  V_last ⇐ V_last * (1 − ζ_a * Y_m)
 3:  V_d ⇐ V_last + (F_in + F_last)/2 * Y_m
 4:  V_d ⇐ max(0, V_d − m_v * 9.8 * μ * Y_m)
 5:  return V_d
```

Figure 5C, produces a linear transition at the boundary of the permitted area, resembling the characteristic curves of a spring.

In our simplified 1D example, we define the Spring Force Map $F_{spr}(x)$ as a mapping function that returns the magnitude of the spring force at position $x$. The Spring Force Map is calculated using a 1-flipped convolution result between the Motion Restriction Map $M_{rs}(x)$ and the impulse convolution kernel:

$$F_{spr}(x) = K_{sp} L_{max}(1 - M_{rs}(x) \circledast Kernal_{sw}(x)) \quad (4)$$

where in Equation 4, $K_{sp}$ denotes the spring stiffness, and $L_{max}$ denotes the width of the impedance force zone. The direction of the spring force at position $x$ $F_{dir}(x)$ is computed using the sign of the negative local derivative:

$$F_{dir}(x) = \text{sgn}\left(-\frac{\partial F_{spr}(x)}{\partial x}\right) \quad (5)$$

Where in Equation 5, the sgn() is the sign function.

The damper force is calculated based on the direction of the spring force and the current end-effector velocity.

### 5.3.2 Impedance force calculation in 2D

The same concept can be generalized into 2D space. The 2D Spring Force Map for any motion restriction map $M_{rs}$ can be generated using a 2D convolution between the Motion Restriction Map and a 2D uniformly filled circular kernel. Similar to the 1D case, the direction of the 2D spring force can be computed using the negative local gradient. The width of the impedance force zone can be controlled by the diameter of the circular kernel.

Figures 5D–G illustrates an example spring force vector map calculation process for a hand-drawn restriction map. The first step is to expand the trajectory in the radial direction. The reason for the expansion is that: In Figure 5B, the permitted area boundaries are at $x = 2$ and $x = 6$, but in Figure 5C, the generated impedance force zone is between $1.5 \leq x \leq 2.5$ and $5.5 \leq x \leq 6.5$. The generated impedance force zone is centered at the original boundary location, with a width equal to the diameter of the convolution kernel. Therefore, to create room for the impedance force zone, the permitted area must be expanded radially by the radius of the convolution kernel. This expansion is achieved through convolution using the same 2D uniformly filled circular kernel, followed by thresholding the resultant output (setting $x = 1$ if $x > 0$, $x = 0$ otherwise for all $x$ in ($M_{rs} \circledast Kernal_{cir}$)). Figure 5E shows the resulting expanded Motion Restriction Map using a 6 mm radius uniformly filled circular kernel. Figure 5F shows the generated Spring Force Map with a 12 mm impedance zone along the trajectory. Figure 5G shows the derived spring force vector field using the proposed impedance force calculation algorithm and the Spring Force Map.

Algorithm 3 shows the implementation of the Impedance Controller. It takes the input force $F_{in}$, the velocity on last time step $V_{last}$, the current position $P_{in}$, the precalculated Spring Force Map $F_{spr}$, the spring stiffness $K_{sp}$, and the damping ratio $ζ_i$ as input. And output the compositional force $F_o$.

## 6 Experiments

The experiments were conducted in three parts: first, a complexity analysis of the proposed high-level control; second, a performance assessment of GARD's various operation modes and control blocks; and third, a comparison to state-of-the-art end-effector upper limb rehabilitation devices. The experiments involved a single healthy human subject (the engineer). This study is exempt from Institutional Review Board (IRB) approval, as it is in the initial design and functionality testing phase, which does not include clinical research or human factors research involving multiple participants. The subject's involvement was limited to engineering assessments, ensuring no risk or ethical concerns that would typically require IRB oversight.

### 6.1 Complexity analysis of control algorithms

The complexity analysis of the proposed control algorithms was conducted with respect to taskspace discretization precision ($n \times n$ taskspace discretization). The results are displayed in Table 1. The proposed control algorithms achieve an overall runtime and space complexity of $O(n)$, which is manageable for any modern PC or MCU.

An important feature highlighted by the complexity analysis is that, in our method, a complex maze-like Motion Restriction Map

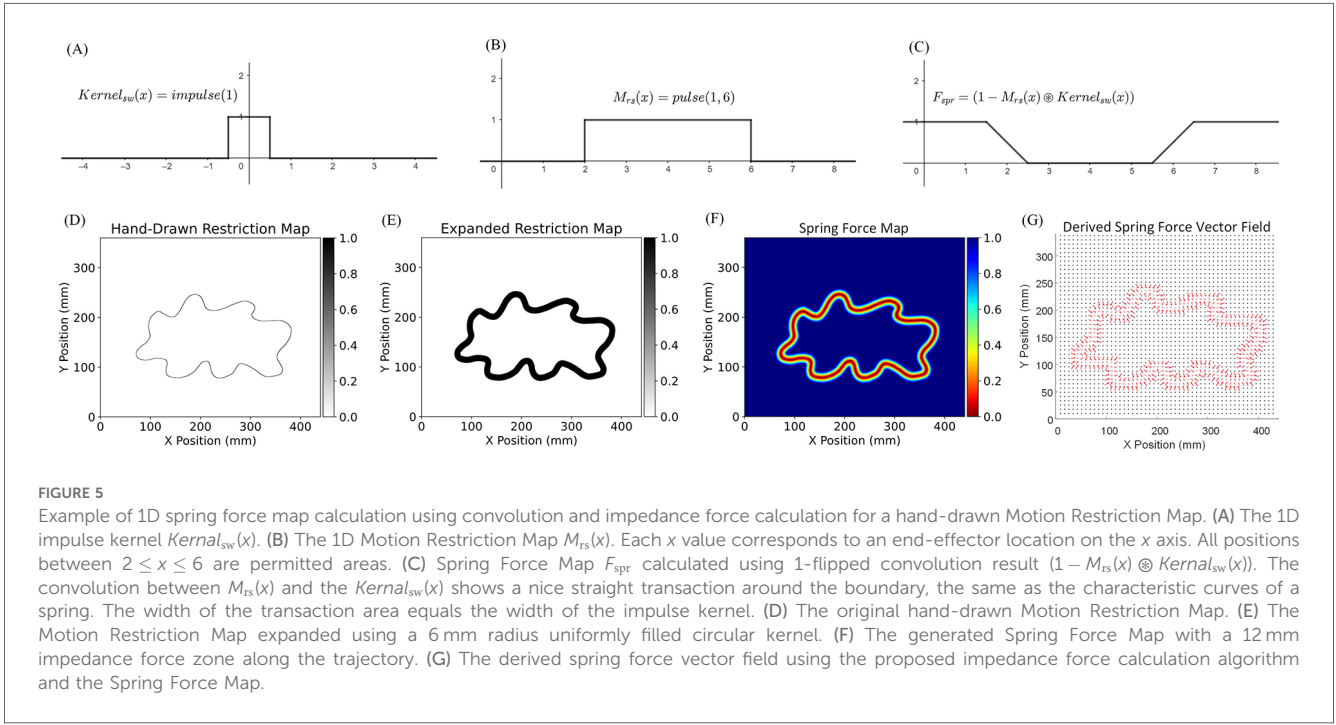

FIGURE 5
Example of 1D spring force map calculation using convolution and impedance force calculation for a hand-drawn Motion Restriction Map. (A) The 1D impulse kernel $Kernel_{sw}(x)$. (B) The 1D Motion Restriction Map $M_{rs}(x)$. Each $x$ value corresponds to an end-effector location on the $x$ axis. All positions between $2 \leq x \leq 6$ are permitted areas. (C) Spring Force Map $F_{spr}$ calculated using 1-flipped convolution result $(1 - M_{rs}(x) \circledast Kernel_{sw}(x))$. The convolution between $M_{rs}(x)$ and the $Kernel_{sw}(x)$ shows a nice straight transaction around the boundary, the same as the characteristic curves of a spring. The width of the transaction area equals the width of the impulse kernel. (D) The original hand-drawn Motion Restriction Map. (E) The Motion Restriction Map expanded using a 6 mm radius uniformly filled circular kernel. (F) The generated Spring Force Map with a 12 mm impedance force zone along the trajectory. (G) The derived spring force vector field using the proposed impedance force calculation algorithm and the Spring Force Map.

**Algorithm 3** Impedance controller.

**Input:** $F_{in}$, $V_{last}$, $P_{in}$, $F_{spr}$, $K_{sp}$, $\zeta_i$,
**Output:** $F_o$
1: $[P_x, P_y] \Leftarrow P_{in}$
2: $F_{dx} \Leftarrow F_{spr}[P_y, P_x + 1] - F_{spr}[P_y, P_x - 1]$
3: $F_{dy} \Leftarrow F_{spr}[P_y + 1, P_x] - F_{spr}[P_y - 1, P_x]$
4: $F_{dir} \Leftarrow normalize(-[F_{dx}, F_{dy}])$
5: $F_o \Leftarrow F_{in} + K_{sp} * F_{spr}[P_{in}] * F_{dir}$
6: $F_o \Leftarrow F_o + \zeta_i * (V_{last} \cdot F_{dir}) * F_{dir}$
7: **return** $F_o$

**Algorithm 4** IEVC helper functions implementation.

1: **procedure** $inCircle(x, y, lineWidth)$
2:   **return** $abs(x^2 + y^2 - 1) < lineWidth$
3: **end procedure**
1: **procedure** $circleMap(R_{circle})$:
2:   $mapLen \Leftarrow R_{circle} * 2 + 10$
3:   $M_{cir} \Leftarrow emptyList()$
4:   **for all** $[x, y]$ in $M_{cir}$ **do**
5:     $dx \Leftarrow x - R_{circle} - 5$
6:     $dy \Leftarrow y - R_{circle} - 5$
7:     $ndx \Leftarrow dx/R_{circle}$
8:     $ndy \Leftarrow dy/R_{circle}$
9:     $wLine \Leftarrow 2/R_{circle}$
10:    **if** $inCircle(ndx, ndy, wLine)$ **then**
11:      $M_{cir}.append([dx, dy])$
12:    **end if**
13:  **end for**
14:  **return** $M_{cir}$
15: **end procedure**
1: **procedure** $findIntersec(M_{rs}, R_{circle}, P_{in})$
2:   $[P_x, P_y] \Leftarrow P_{in}$
3:   $M_{cir} \Leftarrow circleMap(R_{circle})$
4:   $candidatesList \Leftarrow emptyList()$
5:   **for all** $[dx, dy]$ in $M_{cir}$ **do**
6:     **if** $M_{rs}[P_y + dy, P_x + dx] > 0$ **then**
7:       $P_c \Leftarrow [P_x + dx, P_y + dy]$
8:       $candidatesList.append(P_c)$
9:     **end if**
10:  **end for**
11:  **return** $candidatesList$
12: **end procedure**

incurs the same cost as a simple straight-line trajectory Motion Restriction Map. Our method provides a general solution for any 2D motion restriction problem. Moreover, editing a Motion Restriction Map under our method incurs zero runtime computational cost, which indicates that GARD can interact with a dynamically updating motion restriction map in real-time. For example, the therapist can adjust or add new trajectories when the patient is using the GARD, or game developers can dynamically update the level map without interrupting the system. This novel feature holds significant potential for future application development and is a unique capability of our method.

The proposed Assist-As-Needed Mode with soft boundary (the most complex mode), implemented with C#, under a 2200 × 1700 taskspace discretization, achieves an average time step cost of 10 μs when running on an i7-8750H laptop. This demonstrates that solving the 2D Motion Restriction Control problem on modern CPUs using the proposed algorithm leaves a significant computational surplus. Furthermore, the algorithm's high degree of parallelizability underscores its potential for further runtime optimization, suggesting its considerable potential for real-time 3D taskspace motion restriction control.

## 6.2 Experimental validation of powered mode

One of the primary functions of the rehabilitation robot revolves around trajectory tracking. By employing a non-backdrivable end-effector, the user can strictly follow the desired

TABLE 1 Algorithm runtime complexity with respect to taskspace discretization accuracy.

| Function | Time complexity | Space complexity |
|---|---|---|
| IEVC circleMap | $O(1)$ with cache | $O(n)$ |
| IEVC findIntersec | $O(n)$ | $O(n)$ |
| IEVC choose candidate | $O(n)$ | $O(1)$ |
| IEVC overall | $O(n)$ | $O(n)$ |
| Admittance virtual dynamics | $O(1)$ | $O(1)$ |
| Impedance controller | $O(1)$ | $O(1)$ |
| Adding trajectories | 0 | 0 |
| Adding permitted areas | 0 | 0 |
| Adding prohibited areas | 0 | 0 |
| **Proposed method overall (runtime)** | **$O(n)$** | **$O(n)$** |

The bold fonts indicate the outstanding features of this method.

trajectory without interference from external forces in Powered Mode. Figure 6 shows the results of trajectory-following tests in Powered Mode. One healthy subject executed Powered Mode on two equation-based and two hand-drawn trajectories. Both the desired and measured end-effector positions were recorded and plotted. The error was calculated by measuring the closest distance to the permitted area for each recorded end-effector position. The average Mean Absolute Error (MAE) across the four trials between the intended and recorded end-effector positions was 0.012 mm.

## 6.3 Experimental validation of transparent mode

The Transparent Mode has two primary functionalities. First, it incorporates an Admittance Virtual Dynamics control block, which emulates the non-backdrivable end-effector as a free mass in the real world. Second, it enables free movement with a range of motion (ROM) restriction. Two experiments were performed to verify these two functionalities.

The performance of the Admittance Virtual Dynamics simulation is key to achieving smooth and natural motion of the end-effector in both Transparent Mode and Assist-As-Needed Mode. Figures 7A,B depict six trials of circular motion performed in Transparent Mode by a healthy subject. These trials simulated objects with masses of 5, 10, 15, 20, 25, and 30 kilograms moving freely on a plane with a friction coefficient of 0.02. The user was instructed to follow an approximately circular trajectory. In Figure 7C, the desired and measured velocities of the end-effector simulating a 30 kg mass are shown. The overall outcomes are summarized in Table 2. The results indicate that the Admittance Virtual Dynamics block can accurately simulate objects with masses ranging from 10 kg to 30 kg. However, when simulating objects with masses of 5 kg or less, the end-effector experiences larger velocity errors due to the motor acceleration limit.

The range of motion (ROM) is defined as the area bounded by a closed trajectory traversed by the end-effector. The ROM Motion Restriction Map is generated by filling all triangles constructed between the starting position and two consecutive positions of the end-effector. Figures 8A,B demonstrate the generation of a Motion Restriction Map. The trajectory recorded in (A) is used to generate the Motion Restriction Map in (B). Additionally, Figures 8C,D illustrate an experiment using Transparent Mode with the Motion Restriction Map generated in Figure 8B. In Figure 8C, the end-effector trajectory is shown. This experiment includes both movement within the permitted area and interaction with the area boundary. The end-effector can move freely within the permitted area and interact with a hard-smooth-wall-like boundary. Figure 8D highlights instances at points A, B, and C where dedicated forces were applied in an attempt to exceed the ROM restriction, yet the end-effector remained within the ROM.

## 6.4 Experimental validation of Assist-As-Needed modes

The Assist-As-Needed Mode includes two sub-modes: Assist-As-Needed Mode with hard boundary and Assist-As-Needed Mode with soft boundary. In Assist-As-Needed Mode with hard boundary, the user can move the end-effector along a trajectory with a smooth, hard boundary, utilizing the same control method as the Transparent Mode. In Assist-As-Needed Mode with soft boundary, the user can move the end-effector along a trajectory with an impedance force boundary, which provides assistance force pulling the end-effector back toward the trajectory when it deviates. Two key functionalities of the Transparent Modes are highlighted. The first is the motion restriction algorithms applied to the end-effector to ensure the user can only move along the designated trajectory. The second involves impedance control, creating a soft boundary along the trajectory and simulating an impedance spring-damper force that directs the end-effector toward the trajectory. Two experiments were conducted to verify these key functionalities of the Assist-As-Needed Mode.

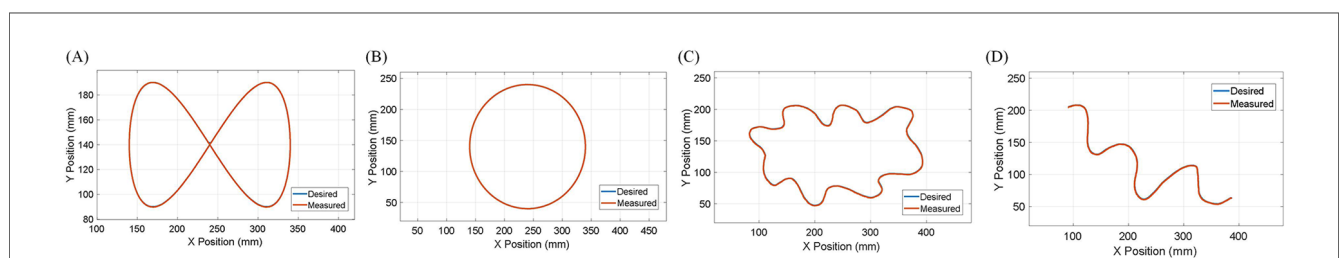

FIGURE 6
Desired and measured end-effector position and error analysis in Powered Mode. One healthy subject operate Powered Mode on two function-based and two hand-drawn trajectories, the desired end-effector positions and measured end-effector positions are recorded and plotted on figures. The error is calculated by measuring the closest distance to the permitted area for each recorded end-effector position if the end-effector is outside the permitted area. (A) Infinite shaped trajectory calculated using equation: $x^4 - x^2 + y^2 = 0$. The Mean Absolute Error (MAE) between desired and measured is 0.00823 mm. (B) Circle shaped trajectory calculated using equation: $x^2 + y^2 - 25^2 = 0$. The MAE between desired and measured is 0.023 mm. (C) Hand drawn circular trajectory. The MAE between desired and measured is 0.00878 mm. (D) Hand drawn line trajectory. The MAE between desired and measured is 0.00763 mm.

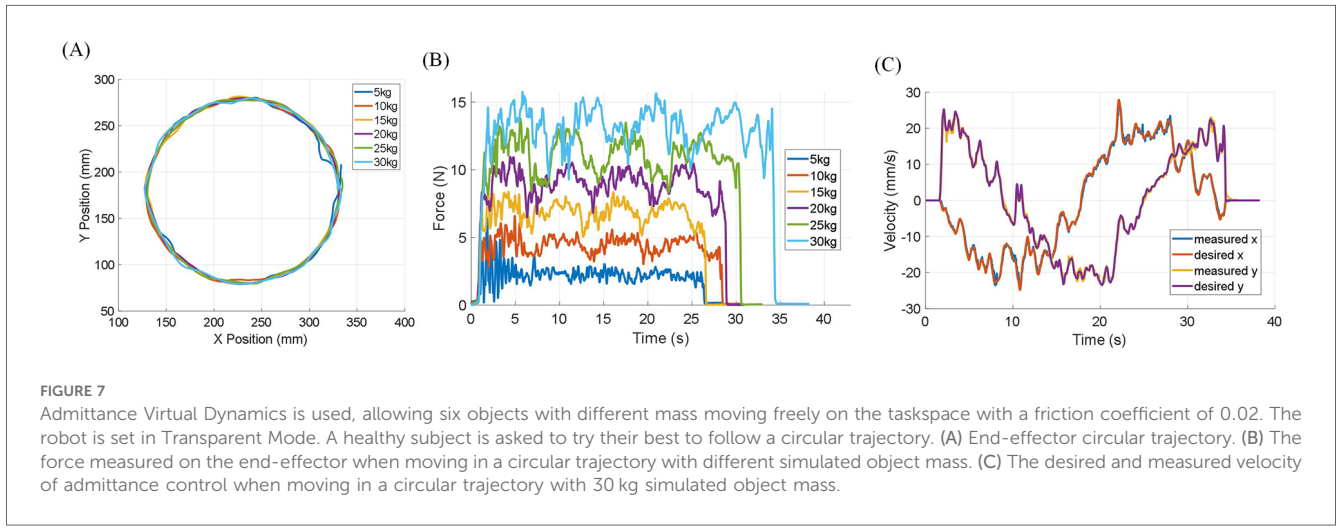

FIGURE 7
Admittance Virtual Dynamics is used, allowing six objects with different mass moving freely on the taskspace with a friction coefficient of 0.02. The robot is set in Transparent Mode. A healthy subject is asked to try their best to follow a circular trajectory. (A) End-effector circular trajectory. (B) The force measured on the end-effector when moving in a circular trajectory with different simulated object mass. (C) The desired and measured velocity of admittance control when moving in a circular trajectory with 30 kg simulated object mass.

TABLE 2 Performance of admittance control simulating object with six different masses moving in circular trajectory under Transparent Mode.

| Mass | 5 kg | 10 kg | 15 kg | 20 kg | 25 kg | 30 kg |
| --- | --- | --- | --- | --- | --- | --- |
| Average measured end-effector force (N) | 2.08 | 3.98 | 6.04 | 8.08 | 9.87 | 11.36 |
| MAE of x-axis desired and measured velocity (mm/s) | 2.45 | 0.75 | 0.41 | 0.46 | 0.37 | 0.35 |
| MAE of y-axis desired and measured velocity (mm/s) | 6.66 | 1 | 0.88 | 0.68 | 0.73 | 0.56 |

The first experiment aims to validate the performance of the IEVC in Motion Restriction Control. A healthy subject was asked to move the end-effector and complete one circular motion per trial. During each trial, a Motion Restriction Map that only allows the end-effector to move along a thin trajectory was loaded into the Motion Restriction Controller. Additionally, a certain level of random noise was injected into the load cell during motion to simulate undesired forces affecting the end-effector's motion. Figure 9 shows a total of six trials of human active trajectory-following tests on two different trajectory shapes, with three different levels of random noise injected into the end-effector. The results show that, in all trials, the recorded trajectory almost completely overlaps with the Motion Restriction

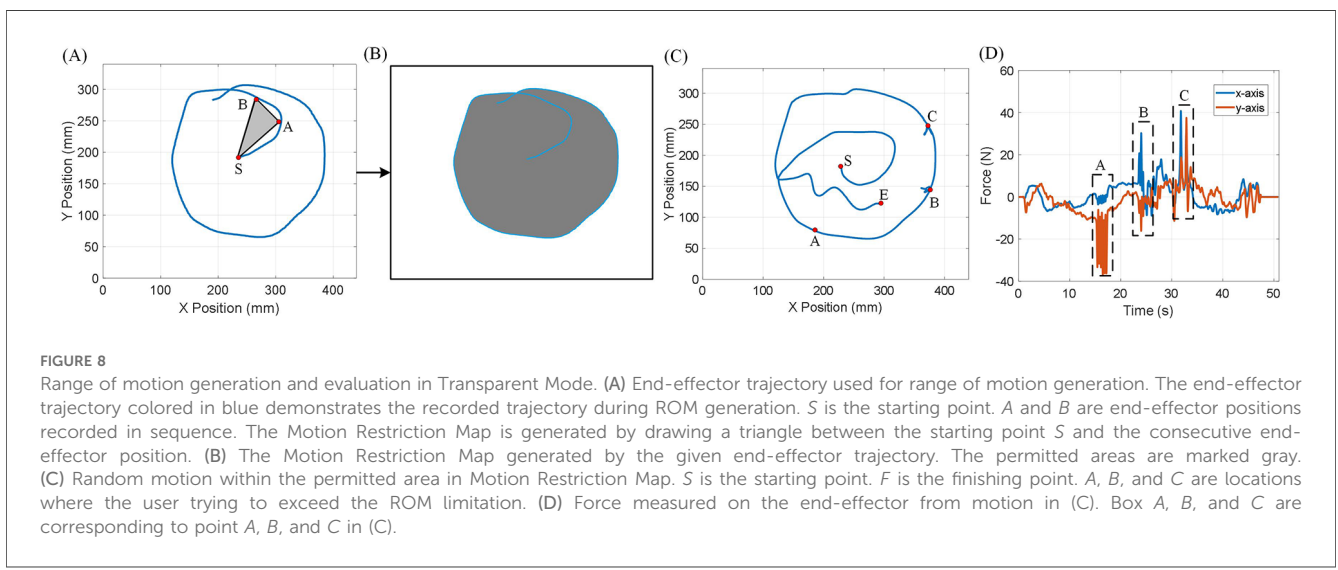

FIGURE 8
Range of motion generation and evaluation in Transparent Mode. (A) End-effector trajectory used for range of motion generation. The end-effector trajectory colored in blue demonstrates the recorded trajectory during ROM generation. S is the starting point. A and B are end-effector positions recorded in sequence. The Motion Restriction Map is generated by drawing a triangle between the starting point S and the consecutive end-effector position. (B) The Motion Restriction Map generated by the given end-effector trajectory. The permitted areas are marked gray. (C) Random motion within the permitted area in Motion Restriction Map. S is the starting point. F is the finishing point. A, B, and C are locations where the user trying to exceed the ROM limitation. (D) Force measured on the end-effector from motion in (C). Box A, B, and C are corresponding to point A, B, and C in (C).

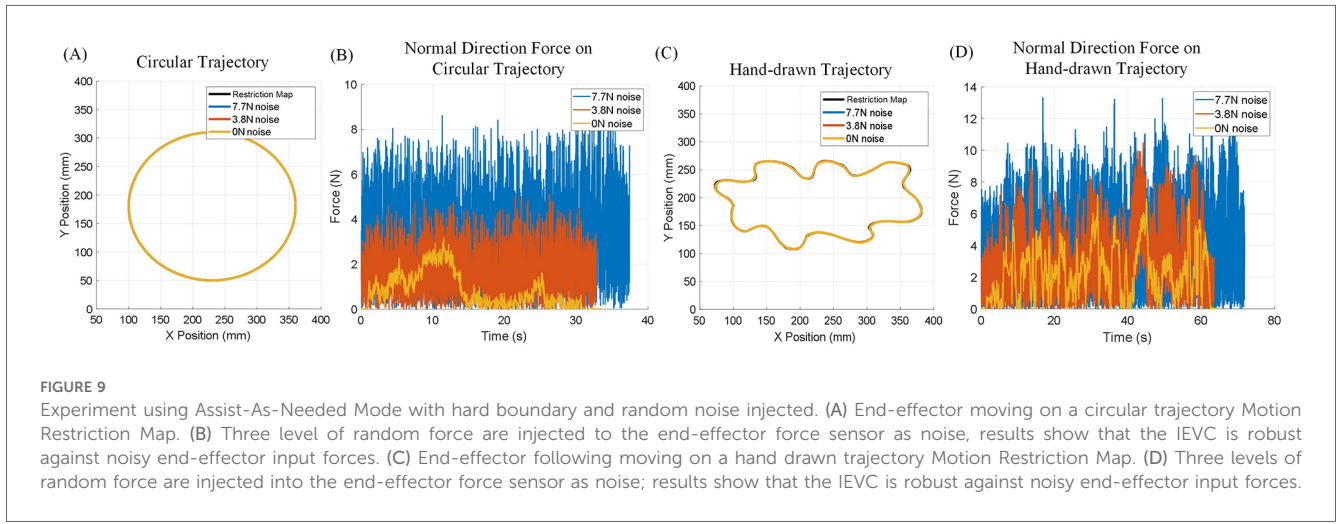

FIGURE 9
Experiment using Assist-As-Needed Mode with hard boundary and random noise injected. (A) End-effector moving on a circular trajectory Motion Restriction Map. (B) Three level of random force are injected to the end-effector force sensor as noise, results show that the IEVC is robust against noisy end-effector input forces. (C) End-effector following moving on a hand drawn trajectory Motion Restriction Map. (D) Three levels of random force are injected into the end-effector force sensor as noise; results show that the IEVC is robust against noisy end-effector input forces.

TABLE 3 Assist-As-Needed Mode with hard boundary and random noise injected.

| Trajectory type | Circular | Circular | Circular | Hand-drawn | Hand-drawn | Hand-drawn |
|---|---|---|---|---|---|---|
| Noise magnitude | 0 N | 3.8 N | 7.7 N | 0 N | 3.8 N | 7.7 N |
| MAE (mm) | 0.067 | 0.061 | 0.032 | 0.134 | 0.115 | 0.113 |

Map, indicating that the proposed control method effectively restricts the end-effector motion on the trajectory. Table 3 presents the quantitative MAE of each trial. The error was calculated by measuring the closest distance to the permitted area for each recorded end-effector position if it is outside the permitted area. Interestingly, the error decreases under higher noise levels. This is because the error in the proposed control algorithm depends only on the moving speed and the trajectory. High levels of injected noise make it difficult for the subject to move the end-effector, thus reducing the moving speed and the error. The results of this experiment demonstrate that the IEVC algorithm can effectively restrict the end-effector motion within any permitted area, making it an ideal choice for the Motion Restriction Controller.

The second experiment involves free movement inside an impedance force zone with a radius of 14 cm. The Spring Force Map is generated using the proposed method with a special Motion Restriction Map, where the permitted area is only at the center of the task space. Figure 10A shows the derived impedance spring force vector field: every permitted location has an impedance spring force pointing toward the center, with a magnitude proportional to the distance from the center. In Figure 10B, one subject was asked to move and hold the end-effector at different locations within the impedance force zone. When the end-effector reached zero acceleration (force equilibrium), the distance between its current position and the center, as well as the current applied force on the end-effector, were recorded and plotted in Figure 10B. The results show that the characteristic curve of the simulated spring area is closely aligned with the characteristic curve of an ideal spring, demonstrating that the proposed impedance control method is effective.

### 6.5 Rehabilitation games with GARD

The GARD user GUI software includes game modes designed to enhance the user experience by making the rehabilitation process more enjoyable and engaging. Currently, two games are implemented: the Breakout game (Figure 11A) and the Maze game (Figure 11B).

The Breakout game is based on the Transparent Mode. In this game, the user controls a paddle to bounce a ball and hit bricks, similar to classic arcade games. The end-effector can move freely across the workspace, with the x-coordinate of the end-effector mapped to the x-coordinate of the paddle.

The Maze game is built on the Assist-As-Needed Mode with Soft Boundary. The objective is to guide an agent through a maze to reach a target position. The end-effector position is mapped to the agent's position, and the Motion Restriction Map corresponds to the maze's walls and paths. In this way, the user can navigate freely inside the maze while interacting with the maze walls, which prevent crossing and allow sliding, mimicking physical walls.

The high degree of freedom in creating custom Motion Restriction Maps and the low time complexity of our control system provide a robust and flexible platform for game development. This flexibility holds great potential for developing new training paradigms.

### 6.6 Comparison to previous researchers

Most previous works include path-following functionality but do not provide quantitative results. Therefore, we perform the comparison qualitatively. Table 4 shows the comparison between our robot and previous research. The Robot-guided path following (impedance) is a trajectory-following function we implemented but did not discuss in previous chapters. This function uses a virtual leading point that automatically marches forward on the trajectory. The virtual leading point serves as the target for impedance control. This function, which is the primary feature of previous works, can be easily achieved by our robot by combining Powered Mode and Admittance Virtual Dynamics.

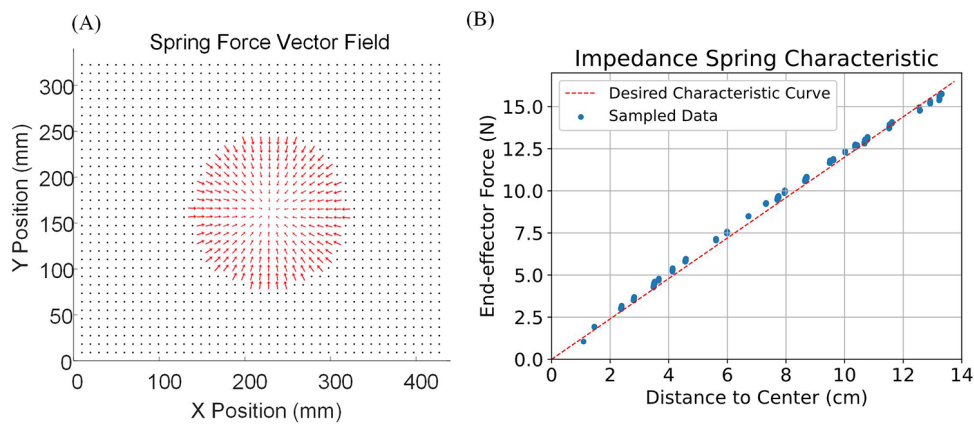

FIGURE 10
Impedance force zone characteristic experiment. (A) The derived impedance spring force vector field. Every permitted location on the map has an impedance spring force pointing towards the center with a magnitude proportional to the distance from the center. (B) One subject is asked to move and hold the end-effector at different locations in the impedance force zone. When the end-effector is at zero acceleration (force equilibrium), the distance between its current position and the center, as well as the current applied force on the end-effector, are recorded and plotted.

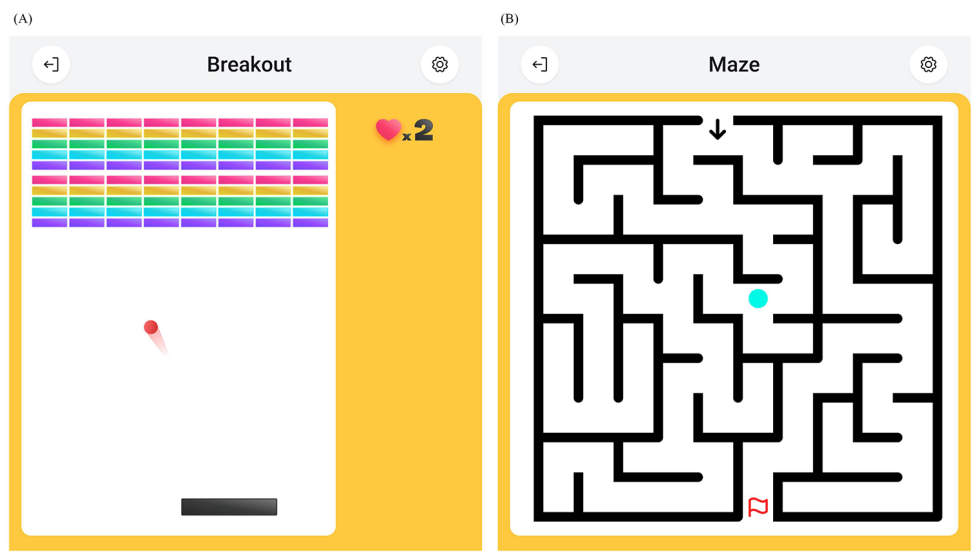

FIGURE 11
Rehabilitation games on GARD GUI software. (A) The Breakout game. A classic arcade game. The user controls a pad to bounce a ball and hit bricks. (B) The Maze game. The user controls an agent to walk through a maze and reach a target position.

TABLE 4 Rehabilitation device comparison.

| Functions | Robot guided path following (impedance) | Robot guided path following (hard) | Impedance boundary trajectory restricted free moving | Impedance boundary area restricted free moving | Hard boundary trajectory/area restricted free moving | Dynamic motion restriction map interaction |
|---|---|---|---|---|---|---|
| HIT Manus (6) | Yes | No | No | No | No | No |
| Physiobot (8) | Yes | No | No | No | No | No |
| H-Man (10) | Yes | No | Yes | No | No | No |
| GARD (ours) | Yes | Yes | Yes | Yes | Yes | Yes |

The proposed control method is the pioneering method that facilitates area-restricted free moving and impedance-boundary free moving while also uniquely supporting dynamic Motion Restriction Map interaction functionality. Additionally, we are the first to deliver quantitative results, establishing a benchmark for future research to reference and compare.

## 7 Conclusion

This study introduces a novel upper limb rehabilitation device, focusing on its design and control methods. GARD uniquely employs a non-backdrivable design with velocity-control-based algorithms, enabling precise, flexible, and stable control during rehabilitation exercises.

One of the key contributions of this work is the development of three control algorithms that address the challenges introduced by the non-backdrivable design. First, the two-level control hierarchy, in conjunction with Admittance Virtual Dynamics, fully capitalizes on the advantages of the non-backdrivable mechanism while mitigating the complexities involved in modeling complex dynamics. Second, the Implicit Euler Velocity Control (IEVC) algorithm effectively solves the error accumulation issue in the velocity-based Motion Restriction Control and extends the idea of 1D trajectory restriction into 2D area restriction. Third, the convolution-based impedance force calculation algorithm provides an efficient, general solution for handling any 2D restrictions.

The integration of the device and control algorithms results in an upper limb rehabilitation device with three distinct operation modes: Powered Mode, Transparent Mode, and Assist-As-Needed Mode. These modes support stroke patients through various stages of recovery. Experimental validation demonstrated the system's effectiveness in accurate end-effector control across various trajectories and configurations and established a detailed performance baseline for future research endeavors. Compared to state-of-the-art devices, the proposed device offers all existing functionalities with superior performance. Additionally, GARD uniquely provides functionalities such as hard boundary area-restricted free moving and dynamic motion restriction map interaction. This device holds strong potential for widespread clinical use, potentially improving rehabilitation outcomes for stroke patients.

## Data availability statement

The original contributions presented in the study are included in the article/Supplementary Material, further inquiries can be directed to the corresponding author.

## Author contributions

FL: Conceptualization, Software, Validation, Visualization, Writing – original draft, Writing – review & editing. YG: Conceptualization, Software, Validation, Visualization, Writing – original draft, Writing – review & editing. WX: Software, Writing – review & editing. WZ: Software, Writing – review & editing. FZ: Supervision, Writing – review & editing. BW: Methodology, Writing – review & editing. HD: Writing – review & editing. CZ: Supervision, Writing – review & editing.

## Funding

The authors declare that no financial support was received for the research, authorship, and/or publication of this article.

## Conflict of interest

FL, YG, WX, WZ, FZ, BW, HD, CZ were employed by Futronics (NA) Corporation.

## Publisher's note

## Supplementary material

The Supplementary Material for this article can be found online at: https://www.frontiersin.org/articles/10.3389/fresc.2024.1469491/full#supplementary-material

## References

1. Virani SS, Alonso A, Aparicio HJ, Benjamin EJ, Bittencourt MS, Callaway CW, et al. Heart disease and stroke statistics-2021 update: a report from the American heart association (2021).

2. Kwakkel G, Kollen BJ, van der Grond J, Prevo AJ. Probability of regaining dexterity in the flaccid upper limb: impact of severity of paresis and time since onset in acute stroke. *Stroke.* (2003) 34:2181–6. doi: 10.1161/01.STR.0000087172.16305.CD

3. Chang WH, Kim Y-H. Robot-assisted therapy in stroke rehabilitation. *J Stroke*. (2013) 15:174. doi: 10.5853/jos.2013.15.3.174

4. Johansen T, Sørensen L, Kolskår KK, Strøm V, Wouda MF. Effectiveness of robot-assisted arm exercise on arm and hand function in stroke survivors-a systematic review and meta-analysis. *J Rehabil Assist Technol Eng*. (2023) 10:20556683231183639. doi: 10.1177/20556683231183639

5. Data from: Inmotion arm for neurological rehabilitation (2024). BIONIK (accessed July 15, 2024).

6. Krebs HI, Ferraro M, Buerger SP, Newbery MJ, Makiyama A, Sandmann M, et al. Rehabilitation robotics: pilot trial of a spatial extension for mit-manus. *J Neuroeng Rehabil*. (2004) 1:1–15. doi: 10.1186/1743-0003-1-5

7. Chua K, Kuah C, Ng C, Yam L, Budhota A, Contu S, et al. Clinical and kinematic evaluation of the h-man arm robot for post-stroke upper limb rehabilitation: preliminary findings of a randomised controlled trial. *Ann Phys Rehabil Med*. (2018) 61:e95. doi: 10.1016/j.rehab.2018.05.203

8. Villar BF, Viñas PF, Turiel JP, Marinero JCF, Gordaliza A. Influence on the user's emotional state of the graphic complexity level in virtual therapies based on a robot-assisted neuro-rehabilitation platform. *Comput Methods Programs Biomed*. (2020) 190:105359. doi: 10.1016/j.cmpb.2020.105359

9. Data from: Armmotus m2 upper limb rehabilitation robotics (2024). Fourier Intelligence (accessed July 15, 2024).

10. Maqsood K, Luo J, Yang C, Ren Q, Li Y. Iterative learning-based path control for robot-assisted upper-limb rehabilitation. *Neural Comput Appl*. (2023) 35:23329–41. doi: 10.1007/s00521-021-06037-z

11. Díaz I, Catalan JM, Badesa FJ, Justo X, Lledo LD, Ugartemendia A, et al. Development of a robotic device for post-stroke home tele-rehabilitation. *Adv Mech Eng*. (2018) 10:1687814017752302. doi: 10.1177/1687814017752302

12. Gandolfi M, Valè N, Dimitrova EK, Mazzoleni S, Battini E, Filippetti M, et al. Effectiveness of robot-assisted upper limb training on spasticity, function and muscle activity in chronic stroke patients treated with botulinum toxin: a randomized single-blinded controlled trial. *Front Neurol*. (2019) 10:41. doi: 10.3389/fneur.2019.00041

13. Lu EC, Wang R, Huq R, Gardner D, Karam P, Zabjek K, et al. Development of a robotic device for upper limb stroke rehabilitation: a user-centered design approach, Paladyn. *J Behav Rob*. (2011) 2:176–84. doi: 10.2478/s13230-012-0009-0

14. Miao Q, Zhang M, Wang Y, Xie SQ. Design and interaction control of a new bilateral upper-limb rehabilitation device. *J Healthc Eng*. (2017) 2017:7640325. doi: 10.1155/2017/7640325

15. Kerr A, Smith M, Reid L, Baillie L. Adoption of stroke rehabilitation technologies by the user community: qualitative study. *JMIR Rehabil Assist Technol*. (2018) 5(2): e9219. doi: 10.2196/rehab.9219

16. Ouendi N, Hubaut R, Pelayo S, Anceaux F, Wallard L. The rehabilitation robot: factors influencing its use, advantages and limitations in clinical rehabilitation. *Disabil Rehabil Assist Technol*. (2024) 19:546–57. doi: 10.1080/17483107.2022.2107095

17. Hussain A, Dailey W, Hughes C, Tommasino P, Budhota A, Gamage WKC, et al. Preliminary feasibility study of the h-man planar robot for quantitative motor assessment. In: *2015 IEEE/RSJ International Conference on Intelligent Robots and Systems (IROS)*. IEEE. p. 6167–72.

18. Maier M, Ballester BR, Verschure PF. Principles of neurorehabilitation after stroke based on motor learning and brain plasticity mechanisms. *Front Syst Neurosci*. (2019) 13:74. doi: 10.3389/fnsys.2019.00074

19. Wade D, Langton-Hewer R, Wood VA, Skilbeck C, Ismail H. The hemiplegic arm after stroke: measurement and recovery. *J Neurol Neurosurg Psychiatry*. (1983) 46:521–4. doi: 10.1136/jnnp.46.6.521

20. Indrawati I, Sajidin M. Active, passive, and active-assistive range of motion (rom) exercise to improve muscle strength in post stroke clients: a systematic review. In: *9th International Nursing Conference, Surabaya, Indonesia*. (2018). Available online at: https://www.scitepress.org/Papers/2018/83248/83248.pdf.

21. Veldema J, Jansen P. Resistance training in stroke rehabilitation: systematic review and meta-analysis. *Clin Rehabil*. (2020) 34:1173–97. doi: 10.1177/0269215520932964

22. Turville ML, Cahill LS, Matyas TA, Blennerhassett JM, Carey LM. The effectiveness of somatosensory retraining for improving sensory function in the arm following stroke: a systematic review. *Clin Rehabil*. (2019) 33:834–46. doi: 10.1177/0269215519829795

23. Hildebrand FB. *Introduction to Numerical Analysis*. New York, NY: Dover (1987). Available online at: https://search.worldcat.org/title/1023961967 (accessed November 20, 2024).